# BADAM: A Public Dataset for Baseline Detection in Arabic-script Manuscripts

Benjamin Kiessling
Alexander von Humboldt-Chair for
Digital Humanities, Leipzig
University
Leipzig, Germany
Université PSL
Paris, France
benjamin.kiessling@psl.eu

Daniel Stökl Ben Ezra
École Pratique des Hautes Études
(EPHE), Université PSL
Paris, France
daniel.stoekl@ephe.psl.eu

Matthew Thomas Miller
Roshan Institute for Persian Studies,
University of Maryland
College Park, Maryland
mtmiller@umd.edu

## ABSTRACT
The application of handwritten text recognition to historical works is highly dependant on accurate text line retrieval. A number of systems utilizing a robust baseline detection paradigm have emerged recently but the advancement of layout analysis methods for challenging scripts is held back by the lack of well-established datasets including works in non-Latin scripts. We present a dataset of 400 annotated document images from different domains and time periods. A short elaboration on the particular challenges posed by handwriting in Arabic script for layout analysis and subsequent processing steps is given. Lastly, we propose a method based on a fully convolutional encoder-decoder network to extract arbitrarily shaped text line images from manuscripts.

## 1 INTRODUCTION

Layout analysis as a major preprocessing step for text recognition is currently considered the limiting factor in the digitization of historical documents both handwritten and printed, especially so for non-Latin writing systems such as Arabic. With the rise of Digital Humanities and large scale institutional digitization projects a significant community of researchers engaged in the improvement of layout analysis on historical material has formed.

The most visible expression of this is a long-standing series of competitions evaluating either layout analysis in isolation [1, 2, 4, 8, 12, 13, 19] or as part of a larger text recognition task such as [3]. Unfortunately, these competitions concern themselves almost exclusively with Western texts written in Latin script despite some efforts to organize competitions on material that is insufficiently treated by current methods.

This euro- and anglocentric focus in document analysis research has changed to some extent recently. Although not directly connected to layout analysis [6] presented binarization, keyword spotting, and isolated character recognition challenges on Balinese palm leaf manuscripts. [7] included a layout analysis task on Arabic manuscripts but notably lacked a publicly available training dataset, except 15 representative images for informational purposes, and participation remained rather modest.

Recognizing that there is an obvious need for a large dataset of non-Western texts we propose a dataset based on one of the most geographically and chronologically extensive manuscript cultures, the Arabic and Persian one. This choice is motivated by multiple reasons: the exceptional size of the available material covering a wide range of topics and styles, complexity of layout rarely encountered in Latin manuscripts, and a large community of scholars working on Arabic-script manuscripts.

In addition, we strive to provide a dataset sufficient in size to support development of state-of-the-art machine learning approaches to layout analysis which despite increasing popularity for Latin documents [5, 10, 20] has seen limited uptake for other writing systems.

### 1.1 Related work

Existing layout analysis datasets capture text lines in a variety of data models. These range from polygons [7, 11, 24], to sub-word bounding boxes [16], down to explicit pixel labeling [12]. Some others such as [1, 3] also include extensive metadata such as reading order, text order, or full transcriptions.

A new paradigm reducing text line segmentation to the successful detection of a continuous sequence of line segments has been established by the ICDAR 2017 Competition on Baseline Detection [8]. There are a plethora of benefits to this minimalistic model: better expression of highly curved baselines in comparison to bounding boxes, lower complexity of training data production than full polygons, easier modelling by semantic segmentation models because of object separability, and the existence of an evaluation scheme [14] that is more directly linked to real world recognition error rates than raw pixel accuracy.

## 2 DATASET

The publicly available and freely licensed BADAM dataset contains 400 annotated scanned page images samples from four digital collections of Arabic and Persian language manuscripts.

### 2.1 Baselines and Arabic Typography

A term arising chiefly from Western typography, the baseline is defined as the virtual line upon which most characters rest with descenders extending below

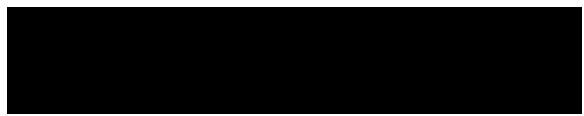
(a) Expulsion of text into the margin

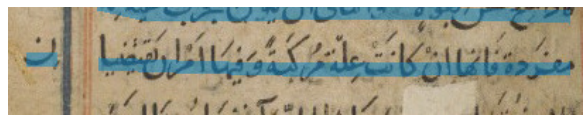
(a) Annotation of dislocated fragments in margin

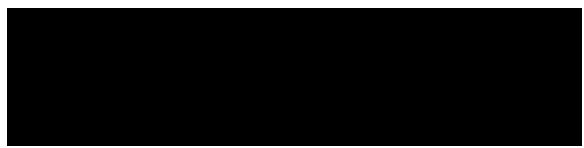
(b) Per-word slanted baselines

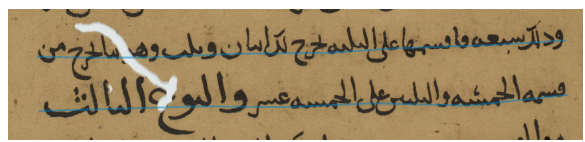
(b) Holes in writing surface

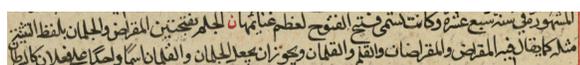
(c) Heaping of words at end of line

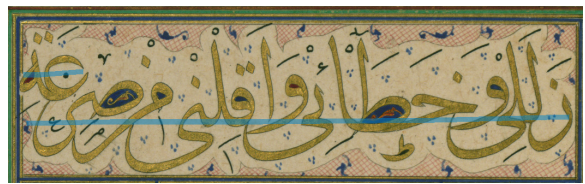
(c) Per-word baseline annotation through imaginary baseline

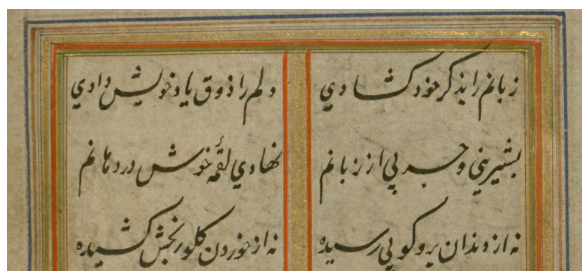
(d) Pseudo-columns in Persian poetry

Figure 1: Aspects of Arabic-script handwriting

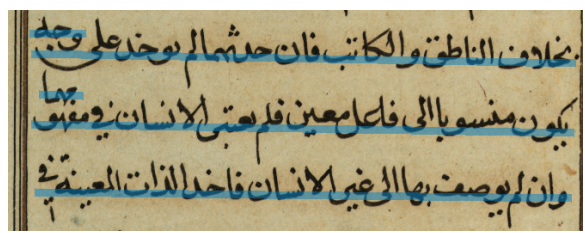
(d) Separate annotation of heaped elements with complete overlap vs single baseline for partial overlap

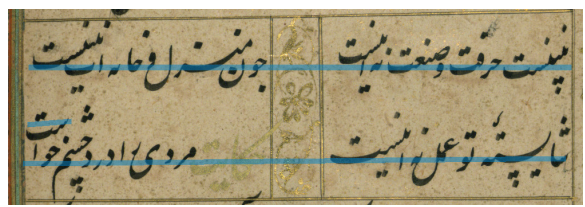
(e) Joint annotation of half-verses as a single baseline

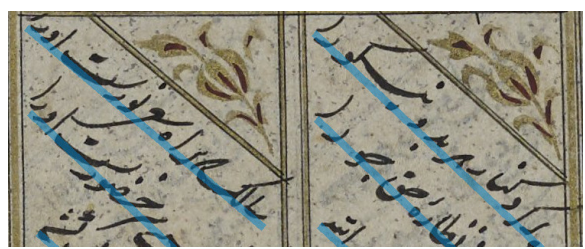
(f) Separated annotation of slanted half-verses

Figure 2: Examples of annotation guideline application (baseline indicated with opaque blue polyline)

While many Arabic handwritten texts present only a single baseline per logical text line a large number of documents, especially calligraphic works in Thuluth and Nastaliq style, display per word slanted baselines (Fig. 1b), multiple baseline levels, and dislocation of fragments into the margins or above other text in the line (heaping) (Fig. 1c and 1a). Most of these cases fulfill the purpose of text justification as hyphenation has been considered unacceptable in Arabic writing for the vast majority of the script's use.

As an additional complication, verses in Arabic poetry almost exclusively consist of two hemistichs, with the half-verse break forming pseudo-columns as shown in Fig. 1d. In some cases there is a combination of pseudo-columns and true multi-column text.

We therefore adopt a modified baseline definition that is oriented towards the current capabilities of text line recognition and reading order determination systems. Text lines are annotated with a single baseline extended through the majority of the line text, except in the cases of majority-overlap heaping (Fig. 2d) and dislocation into the margin (Fig. 2a). In the case of slanted per-word baselines without horizontal overlap a baseline is drawn through an imaginary rotation point at each word (Fig. 2c). A baseline is split in multi-column text and at marginalia/main body boundaires. The hemistichs of poems are annotated as a single baseline per verse (Fig. 2e), except in the case of 45 degree slanted half-verses (Fig. 2f) that cannot easily be connected. In fragmentary material the baseline is continued through faded ink and split at holes in the writing surface (Fig. 2b).

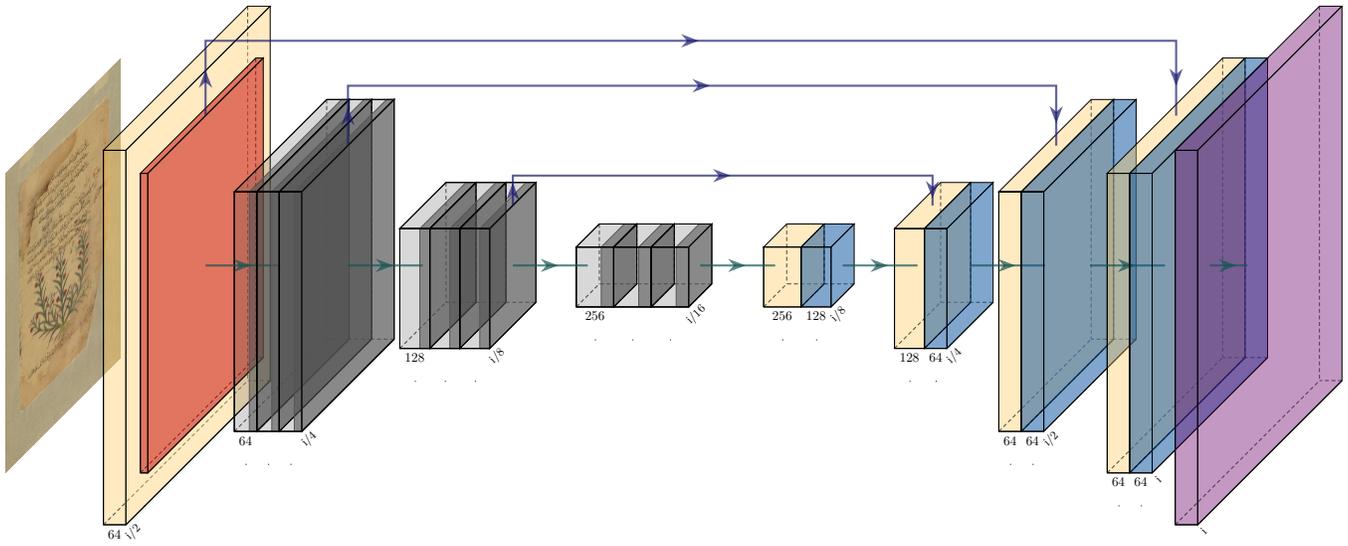

Figure 3: Architecture of the baseline labelling network. Dropout and batch/group normalization layers are omitted. (beige: convolutional layers + ReLU, red: max pooling, grey: ResNet blocks, blue: transposed convolutions, purple: convolution + sigmoid)

These annotation guidelines amount to a conservative estimation of the capabilities of layout analysis systems, specifically their capacity to associate disconnected elements on the page belonging to the same logical line. It is relatively easy to extend the dataset with a more abstract data model that groups multiple baselines into a logical text line and we expect to do so in the future.

## 2.2 Data

42 manuscripts were randomly sampled from the collections of the Qatar Digital Library (15), the digital collection of the Walters Art Museum (13), the Beinecke Rare Book and Manuscript Library (6), and University of Pennsylvania Libraries manuscript collection (8). 10 single page images chosen were annotated for each manuscript with the labelme[1] image annotation tool with the exception of 4 shorter manuscripts from the Beinecke Library containing only 3 to 7 pages. Pages were selected manually for being representative of each work. Overall, there are 10770 lines in the corpus with a range of 3 to 176 lines per manuscript page ($\mu = 30.3, \sigma = 22.1$). The majority of the corpus is written in the Naskh style with the remainder being split between Thuluth, Nastaliq, and Kufic. Other regional styles such as ones used in Ottoman writing are currently absent.

A variety of writings is represented in the corpus:

(1) Medical treatises including poetry with extensive marginalia
(2) Works on logic, commentary on astronomy and arithmetic
(3) Illuminated prayer books and religious texts
(4) Texts on law such as legal glossaries
(5) Illuminated poetic works in Persian and Arabic
(6) Treatises on the legality and rules of chess including extensive diagrams and marginalia

The scan quality of the material varies according to the collection it was sourced from. While all are produced to a professional standard, the resolution varies considerably from 200dpi in the QDL, to 300 dpi in material from the Walters and Beinecke, and 500dpi at the University of Pennsylvania.

A predefined random split into a 320 page training set and a 80 page test set is provided. The annotation is available in both PAGE XML and bit mask image formats. The corpus including both sampled images and ground truth is publicly released under CC-BY-SA 2.0 and available for download on the Zenodo[2] research data archive.

## 3 METHOD

Our method consists of two main stages: a pixel level classification of baselines followed by a lightweight baseline extraction step.

In the first stage a fully convolutional encoder-decoder neural network is used to assign each pixel to a either background or baseline. The second stage is a script- and layout-agnostic postprocessing step operating on the heatmap produced by the neural network. Baselines are vectorized into polylines which are then used to extract rectified rectangular line image suitable for processing by an HTR line recognition system.

### 3.1 Pixel Labeling

The dense pixel-labelling of baselines is performed with a modified U-Net architecture [22]. U-Nets and similar fully convolutional networks [18] are state-of-the-art for general semantic segmentation tasks and have achieved excellent results on the cBad dataset [8].

---

[1] https://github.com/wkentaro/labelme

[2] redacted

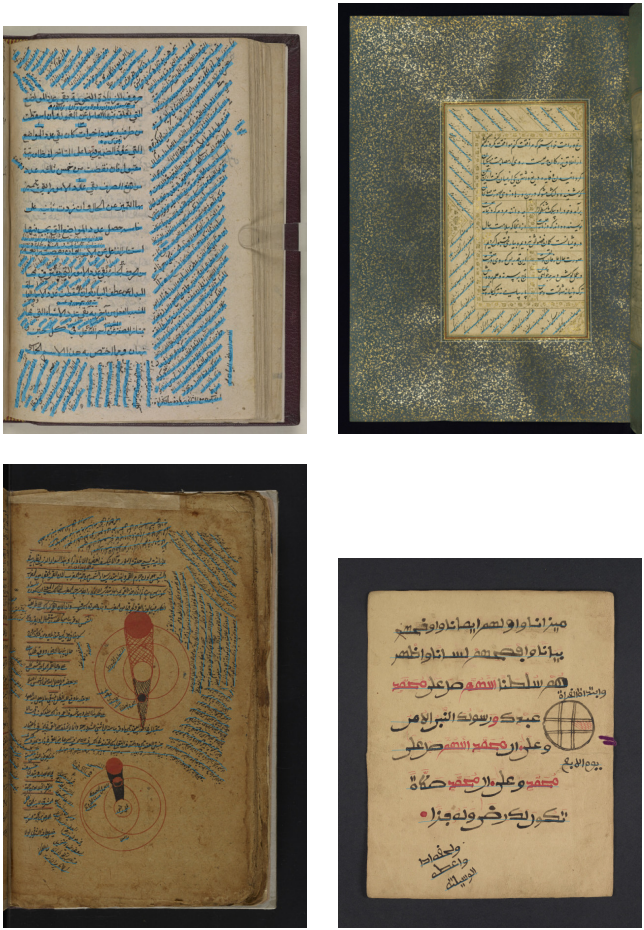

Figure 4: 4 sample pages from the corpus

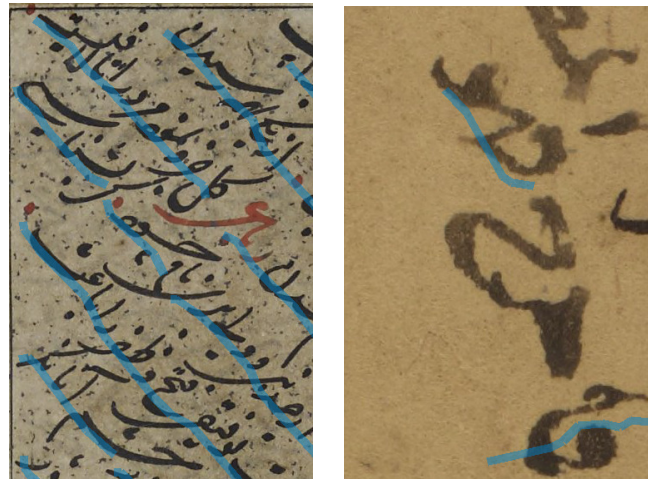

(a) Splitting of calligraphic writing in third/fourth line from top.

(b) Misrecognition of vertical text

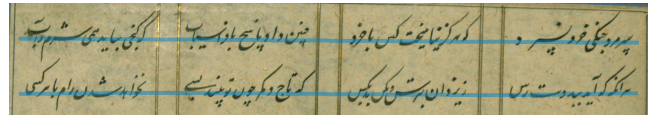

(c) Incorrect splitting of logical 2-column poetry

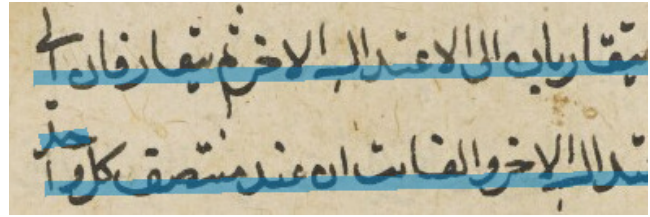

(d) Missed heaped letter in top line, correct example on the bottom

Figure 5: Common error modes of the LA system

The backbone model consists of the first 3 blocks of a 34-layer ResNet in the contracting path followed by 4 3×3 convolution-transposed convolution blocks in the expanding paths with group normalization [25] ($G = 32$) and dropout ($p = 0.1$) employed after each layer and block respectively. A final 1×1 convolutional layer reduces the dimensionality of the input-sized 64-channel feature map to 1, followed by a sigmoid activation. An overall diagram of the network is shown in Fig. 3.

The contracting path is pretrained on ImageNet classification and kept fixed during training of the upsampling blocks. Trainable layers are initialized using the He scheme [15]. We use the Adam optimizer with moderate weight decay ($\alpha = 10^{-4}, \beta_1 = 0.9, \beta_2 = 0.999, w = 10^{-6}$) and early stopping on the binarized F1 score of the validation set. The network is trained on whole color images with the inputs being scaled to a size of 1200 pixels on the shortest edge.

### 3.2 Baseline Estimation

The final sigmoid activation map has to be binarized prior to baseline vectorization. To suppress noise resulting in a higher number of skeleton branches causing a slow down of end point calculation in the next step, the raw heatmap is smoothed with a gaussian filter ($\sigma = 1.5$) first, followed by binarization with hysteresis thresholding ($t_{low} = 0.3, t_{high} = 0.5$)

The binarized image is then skeletonized [17] and 1-connected end point candidates are extracted with a discrete convolution. As the skeleton often contains small branches, determining the actual end points of the centerline skeleton can be challenging. We treat all points along the skeleton as nodes in a graph and assume the true end points are the ones furthest apart on the skeleton. The actual baseline is thus the path of the maximum graph diameter of all possible candidate combinations. This path is then vectorized into a polyline with the Douglas-Peucker algorithm [9].

### 3.3 Line extraction

Vectorized baselines have to be converted into rectangular line images for classification by HTR recognition systems. Given that the baselines found by the system can be highly curved, even circular or spiral-formed, each polyline should be rectified by projecting its line segments and their respective environment consecutively onto a straight baseline.

For each line segment we compute an orthogonal vector of appropriate length including the desired area around the baseline determining the control points above and below the segment at each step. The rasterizations produced by Bresenham's line between both control points at each step are then appended to the rectified line.

According to the results reported in [21] and our own verification on a typeset synthetic dataset the size of the environment extracted around the baseline is not crucial to recognition accuracy as long as the line contents are contained in the rectified line image. We estimate the per-line environment by thresholding the input image with [23], calculating connected components under each baseline, and finding the maximum orthogonal distances of their edges above and below the baseline.

## 4 EVALUATION

We evaluated the proposed method on the 80 page test set using the method described in [14]. The results are shown in table 1. The metrics are slightly lower for our dataset than on the Latin cBAD dataset with a large gap in recall caused by a failure to extract heaped fragments (Fig. 5d) and vertical writing (Fig. ( 5b). On the other hand many missegmented lines are ornate or slanted (Fig. 5a), poetry (Fig. 5c) indicating that the network has not been able to learn a coherent model for these features on the dataset.

The overall agreement in accuracy between the different datasets indicates that modern semantic segmentation methods can be employed for a wide variety of scripts when coupled with appropriate script-agnostic postprocessing. It remains to be seen if the accuracy gap between both datasets can be closed with general purpose systems that are not optimized for a particular set or if script-specific adaptations, such as specialized postprocessing, will be necessary.

|  | P-val | R-val | F-val |
|---|---|---|---|
| **cBAD Simple Track** | | | |
| BYU | 0.878 | 0.907 | 0.892 |
| dhSegment | 0.943 | 0.939 | 0.941 |
| ARU-Net | 0.977 | 0.980 | 0.978 |
| ours | 0.944 | 0.966 | 0.954 |
| **BADAM** | | | |
| ours | 0.941 | 0.901 | 0.924 |

Table 1: Results for the cBAD 2017 dataset and BADAM

## 5 CONCLUSION

We presented a new dataset consisting of 400 annotated page scans of Arabic and Persian manuscripts spanning a wide range of topics and dates of production. Documents in the dataset present various degradations and large differences in the complexity of layout and writing styles. Many of the difficulties posed by them are specific to the Arabic script and should challenge the generalization power of even up-to-date layout analysis methods optimized for Latin script historical documents. While acknowledging that the annotation guidelines oriented on capabilities of current recognition algorithms will likely evolve in the future, our work contributes a solid foundation for comparable evaluation for document analysis researchers.

In addition we describe a baseline system for line extraction from the corpus and evaluate its results, showing that even state-of-the-art methods have difficulties segmenting challenging Arabic handwriting as accurately as Latin manuscripts.